\documentclass[sigconf]{acmart}
\usepackage{tabularx}  
\usepackage{booktabs}  
\usepackage{makecell} 
\usepackage{amsmath}      
\usepackage{algorithm}    
\usepackage{algpseudocode}
\usepackage{enumitem}
\usepackage{colortbl} 
\usepackage{xspace}
\usepackage{array}
\newcolumntype{P}[1]{>{\centering\arraybackslash}p{#1}} 
\newcolumntype{L}[1]{>{\raggedright\arraybackslash}p{#1}} 

\newcommand{\method}{TURA\xspace}
\makeatletter
\newcommand{\thickhline}{%
    \noalign {\ifnum 0=`}\fi \hrule height 1pt
    \futurelet \reserved@a \@xhline
}

\AtBeginDocument{%
  }

\setcopyright{acmcopyright}
\copyrightyear{2018}
\acmYear{2018}
\acmDOI{XXXXXXX.XXXXXXX}

\acmSubmissionID{123-A56-BU3}
\setcopyright{none}
\renewcommand\footnotetextcopyrightpermission[1]{}

\begin{document}

\title{TURA: Tool-Augmented Unified Retrieval Agent for AI Search}

\author{%
Zhejun Zhao$^{*}$, Yuchen Li$^{*}$, Alley Liu$^{*}$, Yuehu Dong, Xiaolong Wei, Lixue Zheng,\\
Pingsheng Liu, Dongdong Shen, Long Xia, Jiashu Zhao, Dawei Yin\\
Baidu Inc.%
}

\renewcommand{\shortauthors}{Zhao et al.}

\begin{abstract}
The advent of Large Language Models (LLMs) is transforming search engines into conversational AI search products, primarily using Retrieval-Augmented Generation (RAG) on web corpora. However, this paradigm has significant industrial limitations. Traditional RAG approaches struggle with real-time needs and structured queries that require accessing dynamically generated content like ticket availability or inventory. Limited to indexing static pages, search engines cannot perform the interactive queries needed for such time-sensitive data. Academic research has focused on optimizing RAG for static content, overlooking complex intents and the need for dynamic sources like databases and real-time APIs. To bridge this gap, we introduce \textbf{TURA} (\textbf{T}ool-Augmented \textbf{U}nified \textbf{R}etrieval \textbf{A}gent for AI Search), a novel three-stage framework that combines RAG with agentic tool-use to access both static content and dynamic, real-time information. TURA has three key components: an Intent-Aware Retrieval module to decompose queries and retrieve information sources encapsulated as Model Context Protocol (MCP) Servers, a DAG-based Task Planner that models task dependencies as a Directed Acyclic Graph (DAG) for optimal parallel execution, and a lightweight Distilled Agent Executor for efficient tool calling. 
Moreover, we release a comprehensive benchmark derived from anonymized Baidu Search production logs with expert annotations, covering diverse static–dynamic query scenarios and standardized evaluation protocols.
TURA is the first architecture to systematically bridge the gap between static RAG and dynamic information sources for a world-class AI search product. Serving tens of millions of users, it leverages an agentic framework to deliver robust, real-time answers while meeting the low-latency demands of a large-scale industrial system.

\end{abstract}





\maketitle
\begingroup
\renewcommand\thefootnote{}
\footnotetext{$^{*}$ Equal contribution.}
\endgroup

\section{Introduction}

\begin{figure}[!t]
    \centering
    \includegraphics[width=\linewidth]{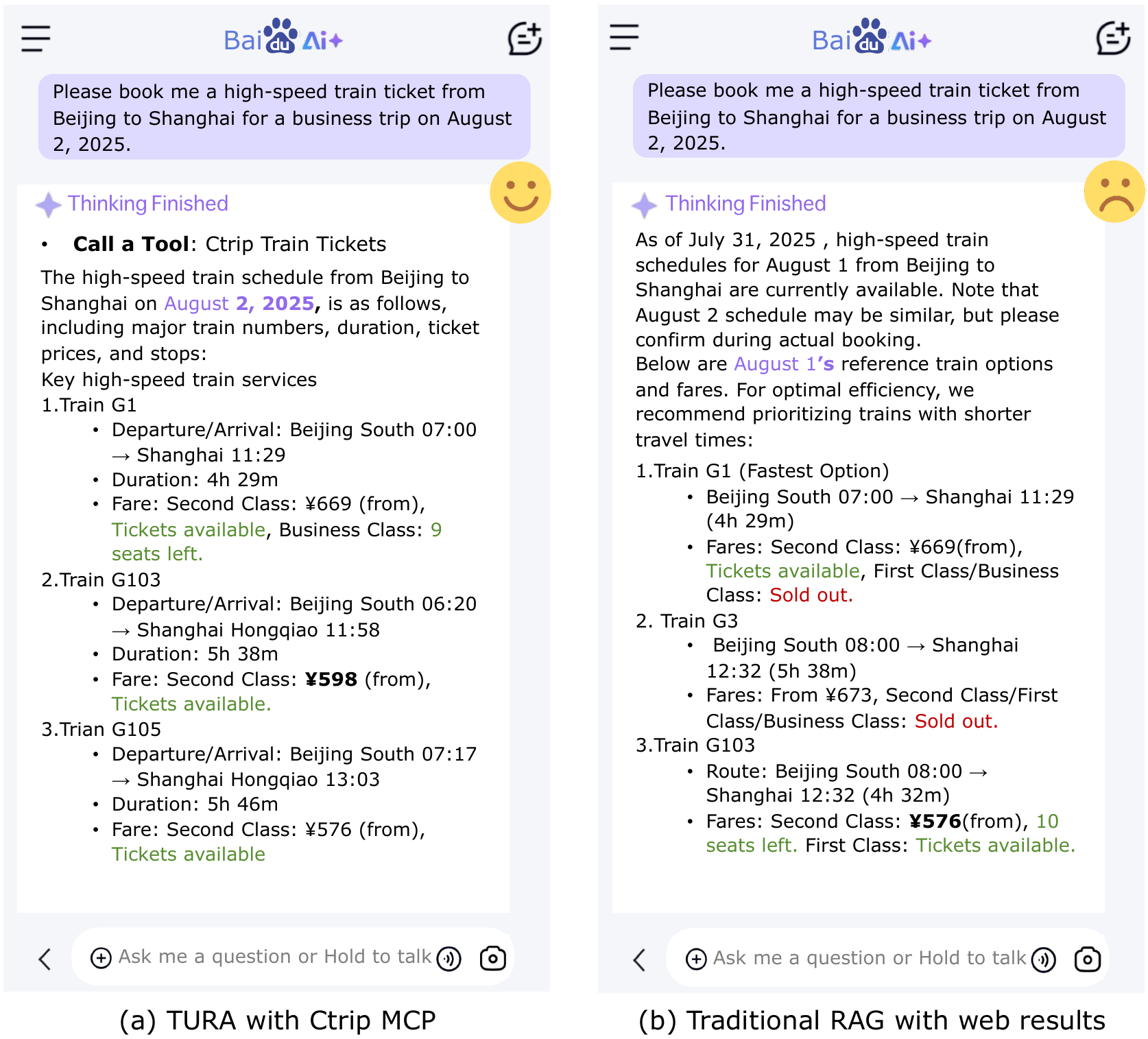}
    \caption{
    Demonstration of TURA's agentic capabilities. Given a query on July 31, 2025: 
    (a) TURA autonomously utilizes a tool by calling Ctrip's API to successfully look up ticket information and complete the required action.
    (b) This contrasts with a traditional RAG-based AI search, which can only retrieve information from static webpages and is unable to perform the required action.
    }
    \vspace{-5mm}
    \Description{comparation}
    \label{fig:product}
\end{figure}

Traditional web search engines, built on foundational algorithms like PageRank \citep{page1999pagerank} and the early Google architecture \citep{brin1998anatomy}, have long relied on the ``ten blue links'' paradigm~\citep{li2025s}. While these systems excel at retrieving and ranking information from vast corpora of unstructured web pages, they have historically struggled with queries demanding structured, real-time, or transactional information, such as flight availability or weather forecasts. To address these limitations, search engines introduced manually curated components like Google's OneBox \citep{elia2017innovative} or Baidu's Aladdin platform \citep{zhang2012envelopment,li2023coltr}, which present information from specific, trusted data sources in dedicated formats. However, these approaches prove fundamentally limited by their reliance on brittle, hand-crafted integrations that are difficult to scale.

The advent of Large Language Models (LLMs) has catalyzed a fundamental paradigm shift in information access \citep{chowdhury2024ai}, transforming search from keyword-based retrieval to conversational, answer-centric systems. The dominant architecture in this new era is Retrieval-Augmented Generation (RAG) \citep{lewis2020retrieval, guu2020retrieval}, which grounds LLM responses in factual knowledge retrieved from web-scale corpora. Commercial deployments like Perplexity AI, ChatGPT's web search integration, and Google's AI Overviews demonstrate the practical viability of this approach \citep{xiong2024search,li2025towards}.

However, this RAG-centric paradigm inherits a critical limitation from its search engine ancestry: it primarily operates on a static snapshot of the web. Current RAG systems, optimized for retrieving from pre-indexed web documents \citep{izacard2020leveraging, xiong2020approximate}, are fundamentally incapable of accessing dynamic, real-time information that is not present on a static webpage but must be generated via interaction. For instance, they cannot query a flight booking system for ticket availability on a specific future date or check real-time inventory from an e-commerce database, as this information is only accessible through interactive queries to APIs or databases. As illustrated in Figure \ref{fig:product}, a standard RAG system fails to answer a time-sensitive query because the necessary information must be generated dynamically, a capability it inherently lacks. This inability to interact with live services renders them ill-equipped for a significant class of user needs that go beyond static information retrieval.

To bridge this critical gap between retrieving from static corpora and interacting with dynamic data sources, we propose \textbf{\method} (\textbf{T}ool-Augmented \textbf{U}nified \textbf{R}etrieval \textbf{A}gent for AI Search), which is a novel three-stage agentic framework that enhances LLMs with the ability to use external tools, moving beyond passive document retrieval to active, real-time data acquisition. The system's design follows the ReAct framework \citep{yao2023react}, which interleaves reasoning and acting, and incorporates advanced planning capabilities inspired by adaptive decomposition methods \citep{prasad2023adapt}. For efficient execution, \method implements DAG-based task decomposition for parallel tool calls \citep{kim2024llm}, addressing latency challenges while maintaining the complex reasoning capabilities demonstrated in large-scale API integration frameworks \citep{qin2023toolllm}. \method's architecture leverages standardized tool interfaces following the Model Context Protocol (MCP) specification \citep{anthropic2024mcp}, enabling seamless integration with diverse information sources through a unified framework. 
Specifically, \method's three-stage architecture comprises the following components:
(1) \textbf{Intent-Aware Tool Retrieval:} This module decomposes complex user queries into atomic sub-intents and performs dense retrieval over a semantically-indexed catalogue of available tools, encapsulating both static document collections and dynamic APIs, to identify relevant candidates for each sub-intent. (2) \textbf{DAG-based Task Planning:} This module constructs optimal and parallelizable execution plans by modeling sub-tasks and their data dependencies as a Directed Acyclic Graph (DAG), enabling the orchestration of complex, multi-hop reasoning chains across multiple tool calls. (3) \textbf{Distilled Agent Executor:} To address critical inference latency barriers in production settings, \method leverages a lightweight yet highly capable agent fine-tuned on curated expert trajectories using a novel mixed-rationale distillation technique, achieving comparable fidelity to proprietary LLMs at a fraction of the computational cost and latency.
We conduct comprehensive offline experiments to rigorously evaluate the performance and efficiency of the proposed \method framework. Offline experimental results reveal that \method substantially improves answer accuracy and faithfulness compared with strong baselines, while comprehensive ablation studies confirm the indispensable role of
each module.
Moreover, since May 2025, \method has been fully deployed, successfully serving tens of millions of users. \method demonstrably expands the capabilities of AI search, providing accurate, real-time answers for a broad spectrum of queries, particularly those involving dynamic, transactional, or non-web data, that were previously intractable for conventional RAG-based systems.
In summary, the contributions of this work could be categorized as follows:
\begin{itemize}[leftmargin=*]
    \item We formulate AI search with heterogeneous information sources as a unified retrieval-and-execution problem over MCP servers, covering both \emph{static} indexed corpora and \emph{dynamic} transactional APIs, and optimize answer quality under strict latency constraints.
    \item We propose TURA, a novel and systematic agentic architecture that effectively integrates diverse, dynamic, and non-web information sources into AI search via tool calling, directly addressing the static-world limitations of traditional RAG systems.
    \item We introduce a cohesive framework of synergistic techniques, including intent-aware tool retrieval, DAG-based task planning, and a latency-optimized distilled agent that collectively solve key industrial challenges of tool selection, task planning, and efficient execution.
    \item 
    We present the first large-scale industrial validation of a tool-augmented agentic search system, demonstrating its viability and effectiveness in a world-class AI search product. In addition, we release MCP-Bench, a comprehensive benchmark constructed from anonymized production logs with expert annotations, providing standardized evaluation protocols and diverse static–dynamic query scenarios to facilitate reproducible research on tool-augmented AI search.
\end{itemize}

\section{Related Work}
\subsection{Retrieval-Augmented Generation}

To mitigate LLM hallucinations and improve factual accuracy \citep{guu2020retrieval,fan2024survey}, RAG systems ground the generation process in external knowledge. These systems typically employ a "retrieve-then-generate" paradigm, with a significant body of research focused on optimizing both retrieval \citep{ma2023query,tao2024llms,lee2025shifting} and generation \citep{liu2023lost,wang2024executable}. Recent advances in RAG have progressed from static retrieve-then-generate pipelines to more dynamic and adaptive frameworks \citep{wang2024corag,fan2024survey}. 
For example, Self-RAG introduces "reflection tokens" that enable the model to decide on-demand whether retrieval is necessary, avoiding the inefficiency of indiscriminate retrieval for every query \citep{asai2024self}. 
Similarly, DRAGIN~\citep{su2024dragin} proposes a dynamic retrieval mechanism that determines both \emph{when} and \emph{what} to retrieve during generation, further improving retrieval scheduling over static pipelines.
In a similar vein, Active Retrieval Augmented Generation adopts an iterative retrieval process throughout generation, allowing the model to gather information as needed. 
Meanwhile, R²AG focuses on bridging the semantic gap between the retriever and the generator by leveraging more nuanced retrieval features \citep{ye2024r2ag}.
Despite these innovations, the practical deployment of RAG in commercial systems such as Perplexity AI and Google AI Overviews continues to reveal challenges in maintaining content quality and robust contextual grounding \citep{williams2024ai}. 
Particularly, most existing RAG frameworks (static or dynamic) primarily assume that all required knowledge can be retrieved from pre-indexed textual corpora. 
Recent studies on inference-time scaling further show that allocating more test-time compute (e.g., iterative prompting or expanded retrieval) can improve answer quality in long-context settings. 
However, these approaches remain fundamentally limited to document retrieval over static indices and do not address queries whose answers must be generated through \emph{live interaction} with external systems such as APIs or databases. 
\emph{In this work, \method targets this complementary regime by enabling tool execution over dynamic, non-web, and transactional data sources while simultaneously optimizing latency through DAG-parallel planning and a distilled executor.}

\subsection{Tool-Augmented Agents}
While RAG systems augment LLMs with textual documents, tool-augmented agents expand their capabilities by providing them with access to a wide array of external resources, such as APIs, web servers, and other computational tools. Much of the research in this area centers on a three-stage process of plan, action, and reflection to guide the agent's behavior \citep{tang2023toolalpaca,xie2024travelplanner}.

The foundational ReAct framework established a paradigm of interleaving reasoning and acting, where the model generates both thought processes and subsequent actions to interact with its environment. This synergy was shown to significantly improve task performance and interpretability. Building on this, Toolformer demonstrated how models could teach themselves to use tools in a self-supervised manner \citep{schick2023toolformer}. The scale of tool integration was dramatically expanded by ToolLLM, which enabled models to leverage thousands of real-world APIs through the use of depth-first search-based decision trees \citep{qin2023toolllm}. Recent work has also focused on optimizing the efficiency of tool use. LLMCompiler, for instance, introduced a Directed Acyclic Graph (DAG)-based approach for parallel tool calling, achieving a 3.6x speedup \citep{kim2024llm}. Agent Q further improved success rates by combining Monte Carlo Tree Search with a self-critique mechanism \citep{putta2024agent}.
Despite these advances, current systems face critical limitations: (1) their workflows are often static and cannot adapt to the complexity of a given query; (2) they struggle with the semantic integration of heterogeneous tools and information sources; and (3) they suffer from coordination inefficiencies, as RAG and tool-augmented systems typically operate in isolation. \emph{In this work, \method is designed to address these issues by proposing a unified architecture that dynamically coordinates retrieval decisions, tool selection, and response generation based on the specific characteristics and intent of the user's query.}

\section{Problem Definition}

Given a user's natural language query $q$, we consider a collection of $N$ heterogeneous MCP servers, $M = \{M_1, M_2, \dots, M_N\}$, each encapsulating a specialized capability that is external to the language model.
These capabilities span both \emph{static information access} (e.g., indexed web search over documents) and \emph{dynamic information generation} (e.g., transactional APIs or database queries).

The core problem is to answer the query $q$ by dynamically composing the functionalities of these MCP servers.
Concretely, the system must autonomously decide \emph{whether to retrieve static web evidence} or
\emph{to execute dynamic tools}, and coordinate their execution to produce a final synthesized answer $A$.
We refer to any concrete composition and ordering of MCP server calls as an execution strategy, denoted by $\pi$.
This formulation emphasizes the \emph{unified} nature of our setting.
Specifically, traditional web retrieval is itself treated as an MCP server (i.e., a search server),
while external APIs and databases are also represented as MCP servers.
As a result, both static retrieval and dynamic tool execution are orchestrated within a single planning-and-execution pipeline and operate over a unified interface, rather than being handled by separate subsystems.

The effectiveness of an execution strategy is governed by a fundamental trade-off between answer quality and system latency.
We aim to maximize the quality of the final answer, $Q(A)$, which depends on the chosen strategy,
while simultaneously satisfying a strict latency constraint, $L(\pi)$.
Our objective is therefore to identify the optimal strategy $\pi^*$ that yields the best possible answer
within a given time budget:
\begin{equation}
    \pi^* = \operatornamewithlimits{arg\,max}_{\pi} Q(A(\pi))
    \quad \text{subject to} \quad
    L(\pi) \leq \tau_{\text{max}},
\end{equation}
where $\tau_{\text{max}}$ denotes the maximum permissible end-to-end latency.
This formulation captures the central challenge of AI search in real-world production settings:
effectively leveraging heterogeneous external capabilities under stringent performance constraints.

\begin{figure*}[htbp]
 \includegraphics[width=0.98\textwidth]{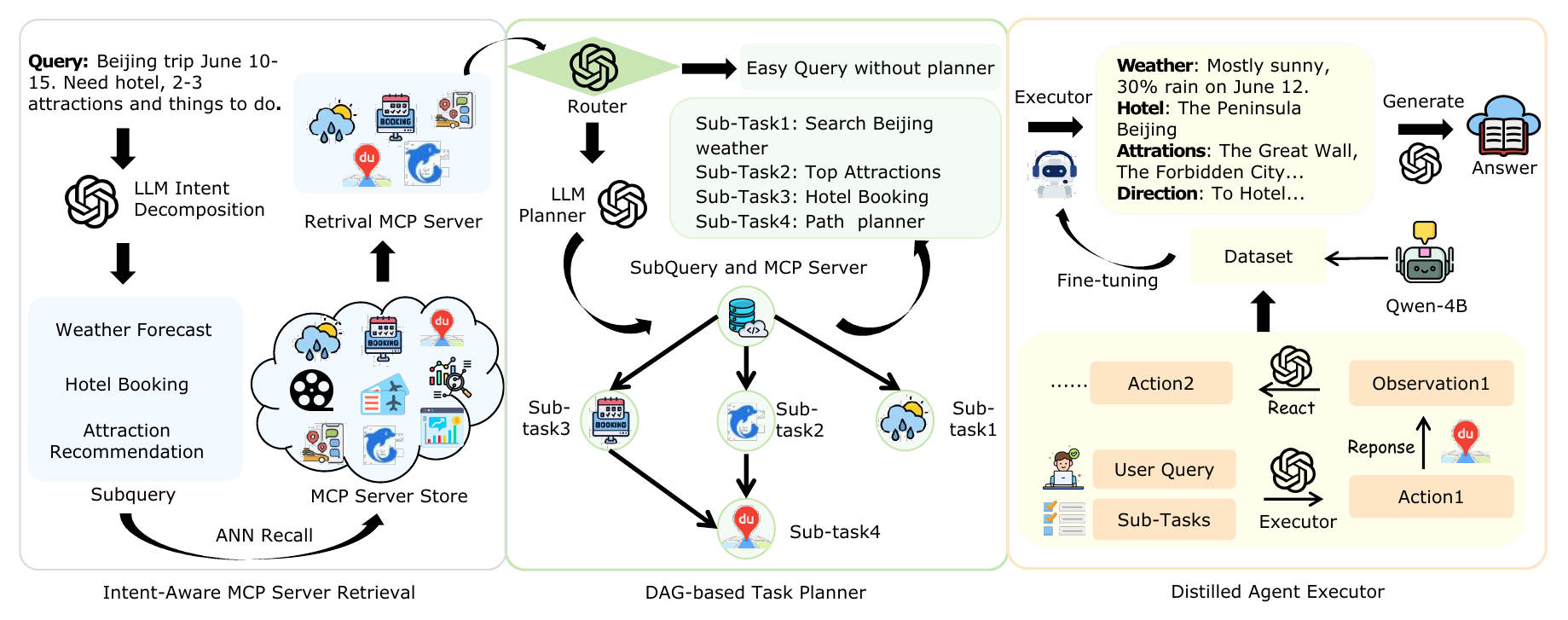}
 \vspace{-6pt}
 \caption{TURA Framework Overview. The framework consists of three stages: Intent-Aware MCP Server Retrieval, DAG-based Task Planner, and Distilled Agent Executor. The example shows the processing progress of \method for the input query ``Beijing trip June 10-15. Need hotel, 2-3 attractions and things to do''.}
 \label{fig:flow_chart}
 \vspace{-3.5mm}
 \Description{TURA three-stage framework architecture}
\end{figure*}

\section{Methodology}
In this section, we first present the overall design of \method, and then introduce the technical details of \method.

\subsection{Framework Overview of \method}
Fig.~\ref{fig:flow_chart} illustrates the proposed framework TURA, which consists of three key modules: \emph{Intent-Aware MCP Server Retrieval Module}, \emph{DAG-based Task Planner Module}, and \emph{Distilled Agent Executor Module}.
Specifically, \method (1) decomposes complex user queries into atomic sub-intents and performs dense retrieval over a semantically-indexed catalogue of available tools, encapsulating both static document collections and dynamic APIs, to identify relevant candidates for each sub-intent. Then, \method constructs optimal and parallelizable execution plans by modeling sub-tasks and their data dependencies as a DAG, enabling the orchestration of complex, multi-hop reasoning chains across multiple tool calls. Eventually, to address critical inference latency barriers in production settings, \method leverages a lightweight yet highly capable agent fine-tuned on curated expert trajectories using a novel mixed-rationale distillation technique to achieve comparable fidelity to proprietary LLMs at a fraction of the computational cost and latency.

\subsection{Intent-Aware MCP Server Retrieval}
This initial stage acts as a filter, efficiently identifying a small, high-recall set of MCP Servers from the global pool $\mathcal{M}$ that are most likely to contribute to answering the query $q$. This prevents the downstream modules from being overwhelmed with irrelevant options.

\subsubsection{\textbf{LLM-based Multi-Intent Query Decomposition}}
User queries are often underspecified and multifaceted. A monolithic query may contain multiple, logically distinct information needs. To handle this, we employ a powerful LLM, $f_{\text{LLM-de}}$, configured with a specific prompt (shown in Appendix~\ref{app:query_decomp}) to function as a query decomposer. The LLM is instructed to parse the raw query $q$ and transform it into a structured set of atomic sub-queries $\mathcal{SQ} = \{sq_1, sq_2, \dots, sq_k\}$, which could be formulated as
\begin{equation}
    \mathcal{SQ} = f_{\text{LLM-de}}(q).
\end{equation}
Each sub-query $sq_j \in \mathcal{SQ}$ is designed to be unambiguous and correspond to a single semantic intent (e.g., ``find the city where the Forbidden City is located.'', ``get weather forecast for a given city''). This decomposition transforms an ambiguous problem into a set of well-defined, tractable sub-problems. This strategy is inspired by recent advances in task decomposition for complex reasoning \citep{zhou2022least}.

\subsubsection{\textbf{Server-level Semantic Index Augmentation}}
A significant challenge in tool use is the ``lexical gap'' between user vernacular and formal API or server descriptions. To bridge this, we perform an extensive offline index augmentation process. For each server $M_i \in \mathcal{M}$, we first define its holistic description $D_i$. Then, we utilize a generative LLM, $g_{\text{LLM-gen}}$ with a specific prompt (shown in Appendix~\ref{app:tool_gen}), to produce a large, diverse set of synthetic queries, $\mathcal{Q}^{\text{syn}}_{i}$, that a user might plausibly issue to access the functionalities of $M_i$. The process can be formulated as
\begin{equation}
    \mathcal{Q}^{\text{syn}}_{i} = g_{\text{LLM-gen}}(D_i, \text{temperature}=T_{\text{high}}).
\end{equation}
By employing a high sampling temperature ($T_{\text{high}} > 1.0$), we encourage the model to explore the periphery of the server's semantic space, generating queries with diverse phrasing and implied intent. This approach, where a model generates training or index data for retrieval, has proven effective in bridging the semantic gap \citep{chen2024re}. The final retrievable unit for each server becomes an augmented document, which is a set of texts $D'_{i} = \{D_i\} \cup \mathcal{Q}^{\text{syn}}_{i}$. This enriched representation provides a dense, multi-faceted semantic footprint for each server.

\subsubsection{\textbf{Dense Vector Retrieval}}
We employ a dense retrieval approach using multi-vector embeddings. During the offline indexing phase, for each server $M_i$, we compute embeddings for all text segments $t \in D'_i$, resulting in a set of vectors $\mathcal{V}_i = \{E(t) \mid t \in D'_i\}$ using a fine-tuned ERNIE \citep{sun2021ernie} model $E(\cdot)$.

During online retrieval, for each sub-query $sq_j$, we embed it as $v_{sq_j} = E(sq_j)$ and perform an Approximate Nearest Neighbor (ANN) search against all server embeddings. The relevance score between a sub-query and a server is determined by the maximum similarity over all of the server's vectors:
\begin{equation}
    \text{sim}(sq_j, M_i) = \max_{v_k \in \mathcal{V}_i} \cos(v_{sq_j}, v_k)
\end{equation}
A MaxSim operation allows for a fine-grained matching between the query's intent and specific facets of the server's functionality.

\subsubsection{\textbf{Multi-Query Score Aggregation}}
The decomposition yields multiple sub-queries, so we must aggregate their retrieval results. For each sub-query $sq_j$, we retrieve a set $\mathcal{P}_j$ of top-ranked (server, score) pairs. We first collect all retrieved pairs from all sub-queries into a single candidate pool:
\begin{equation}
    \mathcal{P}_{\text{cand}} = \bigcup_{j=1}^{k} \mathcal{P}_j
\end{equation}
This pool, $\mathcal{P}_{\text{cand}}$, contains all unique server-score pairs retrieved across all sub-queries. We then employ a maximum score aggregation strategy. For each unique server $m$ that appears in the candidate pool, its final aggregated score, $\text{score}(m)$, is its highest similarity score across all sub-queries:
\begin{equation}
    \text{score}(m) = \max \{ s \mid (m, s) \in \mathcal{P}_{\text{cand}} \}
\end{equation}
This approach ensures that servers demonstrating strong relevance to at least one sub-query are prioritized. Finally, we produce the final retrieved set, $\mathcal{M}_{\text{final}}$, by selecting the top-$K$ servers from the unique servers in $\mathcal{P}_{\text{cand}}$ based on their aggregated score $\text{score}(m)$. This set $\mathcal{M}_{\text{final}}$ serves as the high-recall input for the subsequent DAG-based Task Planner.

\subsection{DAG-based Task Planner}
The planner receives the query $q$, sub-queries $\mathcal{SQ}$, and retrieved servers $\mathcal{M}_{\text{final}}$. A router model then determines the query's complexity. For queries classified as simple, a \emph{single-task execution plan} is constructed without using the DAG planner. For those deemed complex, a dedicated DAG planner is invoked to generate a more sophisticated plan. This acknowledges that for complex reasoning, linear execution plans are often suboptimal, motivating the exploration of non-linear structures like graphs or trees \citep{besta2024topologies}.

The planner, implemented with a highly capable LLM $p_{\text{LLM-plan}}$, is prompted to act as a solution architect with a specific prompt (shown in Appendix~\ref{app:plan}). It analyzes the relationships between sub-queries and the capabilities of the retrieved servers to construct a DAG, $\mathcal{G} = (\mathcal{V}, \mathcal{E})$.

\textbf{Vertices} $\mathcal{V}$: Each vertex $v_k \in \mathcal{V}$ represents a high-level sub-task $st_k$. The planner defines each sub-task as a tuple $st_k = (sq'_k, M_k)$, where $M_k \in \mathcal{M}_{\text{final}}$ is the optimally chosen MCP Server, and $sq'_k$ is a refined, context-aware sub-query. This sub-query might be a direct pass-through of some $sq_j \in \mathcal{SQ}$, or it could be a newly formulated instruction that incorporates the expected output from a parent node in the DAG.

\textbf{Edges} $\mathcal{E}$: A directed edge $(v_a, v_b) \in \mathcal{E}$ indicates a strict data dependency. The planner establishes this edge if the sub-task $st_b$ requires the output of sub-task $st_a$ as part of its input. For example, for the query "Beijing trip June 10-15. Need hotel, 2-3 attractions and things to do.", the planner identifies that Sub-task4 (Path planner) depends on the outputs of Sub-task2 (Top Attractions) and Sub-task3 (Hotel Booking). Specifically, the path planner needs the list of attraction locations and the hotel's address to generate an optimal travel route. Therefore, the planner establishes directed edges $(v_2, v_4)$ and $(v_3, v_4)$ to pass the attraction and hotel information to the path planner sub-task. Meanwhile, Sub-task1 (Search Beijing weather) can be executed in parallel as it has no dependencies.

The output is a structured representation of this DAG, which enables an execution engine to identify and run independent tasks in parallel, drastically reducing the latency $\mathcal{L}(\pi)$ for complex, multi-hop queries.
\subsection{Distilled Agent Executor}
The final stage is the execution of the plan $\mathcal{G}$. An orchestrator traverses the DAG, dispatching each sub-task $st_k = (sq'_k, M_k)$ to our lightweight Agent Executor, $\mathcal{A}_{\theta}$, as it becomes executable. For single-task plans, this process is simplified to a direct execution of the task without DAG traversal. The agent's responsibility is to achieve the goal defined by $sq'_k$ by interacting exclusively with the tools $\mathcal{T}_k$ within the assigned server $M_k$.

Directly using a large-scale LLM like Deepseek-V3 \citep{liu2024deepseek} for this fine-grained execution is infeasible in a real-time system, as the context for each decision (conversation history, server description, dozens of tool APIs) would lead to unacceptable inference latency. We overcome this via agent distillation \citep{kang2025distilling}.

\subsubsection{\textbf{Trajectory Synthesis and Data Curation}}
We first bootstrap a dataset of expert demonstrations, $D_{\text{expert}}$\citep{yin2025magnet}. For a large set of representative sub-tasks, we use a powerful teacher model like Deepseek-V3 to generate execution trajectories with specific prompts (see Appendix~\ref{app:executor}). Each trajectory $\pi_i$ is a sequence of ReAct-style tuples $\langle o_t, \text{th}_t, a_t \rangle_t$, where $o_t$ is the observation, $\text{th}_t$ is the chain-of-thought reasoning, and $a_t$ is the chosen action (a specific tool call within the server).

This raw data is then subjected to a rigorous, automated curation pipeline. The first stage, Correctness Filtering, employs a judge model, $J_{\text{correct}}$\citep{huang2024empirical}, to validate each step of a trajectory. This judge scrutinizes for adherence to API schemas, validity of parameter values, and logical soundness of the thought process leading to the action. Any trajectory failing these checks is discarded.

Subsequently, the second stage, Efficiency Filtering, uses another judge, $J_{\text{efficient}}$, to analyze the now-correct trajectories for performance. It identifies and flags issues such as action redundancy and path sub-optimality
These inefficient trajectories are then either pruned or programmatically corrected. This two-stage curation transforms the noisy expert data into a high-quality, optimal distillation dataset $D_{\text{distill}}$:
\begin{equation}
    D_{\text{distill}} = J_{\text{efficient}}(J_{\text{correct}}(D_{\text{expert}})).
\end{equation}
\subsubsection{\textbf{Mixed-Rationale Supervised Fine-Tuning}}
To achieve minimal inference latency, we fine-tune Qwen3 \citep{yang2025qwen3} series, a much smaller model than large-scale LLMs like Deepseek-V3 on $\mathcal{D}_{\text{distill}}$ using a mixed-rationale supervised fine-tuning (SFT) strategy. The training process explicitly leverages the chain-of-thought data. The agent $\mathcal{A}_{\theta}$ is trained to predict the full sequence of tokens, including both the thought and the action. The loss function is the standard cross-entropy loss over the target sequence:
\begin{equation}
\mathcal{L}_{\text{SFT}}(\theta) = -\sum_{(st', \pi') \in \mathcal{D}_{\text{distill}}} \sum_{t=1}^{|\pi'|} \log P_{\theta}(y_t | y_{<t}, st'),
\end{equation}
where $y_t$ are the tokens of the concatenated thought and action at each step. By training on rationales, the model learns the underlying reasoning process that maps observations to optimal actions.

Critically, during online inference, we provide the agent $\mathcal{A}_{\theta}$ with a specialized prompt that instructs it to directly generate the action, omitting the thought step. Having implicitly learned the reasoning patterns, the model can produce the correct action without the costly auto-regressive generation of the rationale text. This ``train-with-thought, infer-without-thought'' paradigm allows the agent to retain the high-quality decision-making of the teacher model while operating at a fraction of the computational cost and latency.

\section{Evaluation}
In this section, we conduct a series of comprehensive experiments to rigorously evaluate the performance and efficiency of our proposed TURA framework. Our evaluation is structured around four key research questions (RQs):

\begin{itemize}
    \item[\textbf{RQ1:}] How does TURA perform in end-to-end scenarios compared to the baselines, both in offline benchmarks and live online production environments?
    \item[\textbf{RQ2:}] What is the contribution of our proposed Intent-Aware MCP Server Retrieval module? Specifically, how do query decomposition and index augmentation impact retrieval efficacy?
    \item[\textbf{RQ3:}] To what extent does the DAG-based Task Planner improve the system's performance, particularly for complex queries requiring multi-tool coordination?
    \item[\textbf{RQ4:}] How effective is our agent distillation strategy in creating a lightweight yet highly capable executor? Can a smaller language model, when properly distilled, match the execution quality of a much larger teacher model while satisfying strict latency constraints?
\end{itemize}

\subsection{Experimental Setup}

\subsubsection{\textbf{Datasets and Benchmarks}}
To evaluate real-world performance, we built \textbf{MCP-Bench}, a comprehensive benchmark from anonymized production logs that captures natural query distributions from simple lookups to complex multi-hop requests. Working with Baidu's annotation team, we used a rigorous multi-stage protocol where experts annotated each query's ground-truth MCP Servers, execution trajectories, and ideal answers. Cross-validation with multiple annotators and consensus resolution achieved a Cohen's kappa of 0.87 for reliability.
MCP-Bench is constructed from anonymized production logs with a multi-stage expert annotation protocol. We report key statistics to support reproducibility:
\begin{itemize}[leftmargin=*]
  \item \textbf{Queries}: \textit{10,683} total queries, with \textit{95.02\%} single-intent and \textit{4.98\%} multi-intent queries.
  \item \textbf{MCP servers}: \textit{61} servers across \textit{10} categories, including travel, weather, food, finance, and general-purpose ``other'' categories.
  \item \textbf{Tool-call complexity}: average \textit{1.533} tool calls per query (P50=\textit{1}, P90=\textit{3}), and average DAG width \textit{1.147}.
  \item \textbf{Domain distribution.} In addition to search, which accounts for the largest proportion, the remaining top categories by tool-call proportion are travel (29.60\%), finance (11.42\%), weather (7.70\%), food (7.22\%), and other (28.57\%).
  \item \textbf{Evaluation protocol.} We release annotation prompts, judge rubrics, and aggregation scripts to facilitate reproducibility, while anonymizing server and tool identifiers to preserve privacy. 
\end{itemize}
\noindent MCP-Bench is publicly released at \url{https://github.com/weixiaolong94-hub/TURA}.

\subsubsection{\textbf{Baselines}}
We compare TURA against three categories of strong baselines and two ablated versions of our own system:
\begin{itemize}[leftmargin=*]
    \item \textbf{LLM + RAG:} A powerful LLM (Deepseek-V3) combined with a standard RAG pipeline. Its retriever is a specialized variant of the Baidu Search API, which bypasses the reranking stage to provide raw documents. The LLM synthesizes an answer based on the retrieved web content without actively executing tools.
    \item \textbf{Dynamic RAG} adapts retrieval decisions during generation (e.g., when/what to retrieve) but remains constrained to a static index; it cannot interact with transactional APIs.
    \item \textbf{Tool-Agent} (ReAct-style agents and DAG-enabled agents) can call tools, but do not incorporate our unified MCP server retrieval module or our distilled low-latency executor; we run them with the same MCP servers to isolate orchestration and execution differences.
\end{itemize}

\subsubsection{\textbf{Evaluation Metrics}}
To evaluate the performance of \method and baselines, we take a multi-faceted evaluation strategy.
\begin{itemize}[leftmargin=*]
    \item \textbf{End-to-End Offline Evaluation:} We measure \textbf{Answer Accuracy} and \textbf{Faithfulness}. Answer Accuracy assesses whether the final generated answer correctly addresses the user's query. Faithfulness evaluates whether the answer is grounded in and consistent with the information returned by the invoked tools or web pages. Both metrics are evaluated using a combination of human annotation and LLM-as-a-judge on a 3-point scale (Correct/Partially Correct/Incorrect).
    \item \textbf{Online A/B Testing:} In the live production environment, we track standard industry metrics: \textbf{Session Success Rate (SSR)} \citep{deng2016data}, which measures the fraction of user sessions where a satisfactory answer is provided, and \textbf{Good vs. Same vs. Bad (GSB)} \citep{li2025rankexpert}, a human-rated comparison of TURA's output against the production baseline.
    \item \textbf{Component-wise Evaluation:} For detailed ablation studies, we use targeted metrics. For the retrieval module (RQ2), we report \textbf{Recall@5} and \textbf{Precision@5}. For the agent executor (RQ4), we measure \textbf{MCP-Tool Calling Accuracy} and \textbf{Average Latency per Step}.
\end{itemize}

\subsubsection{\textbf{Implementation Details}}
Our TURA implementation utilizes Qwen3-1.7B for query decomposition and ERNIE as the dense retrieval encoder. The DAG planner is implemented using Deepseek-V3. The agent distillation process employs Deepseek-V3 as the teacher model. The resulting student agents are fine-tuned from the Qwen3 series. For latency evaluation, 80th percentile measurements were conducted for tool execution processes. The Qwen3 series models were benchmarked on two NVIDIA L20 GPUs configured identically to the production deployment environment, while Deepseek-V3 was evaluated using the online service hosted on the Baidu Qianfan platform.

\subsection{Overall Performance Evaluation (RQ1)}

\subsubsection{\textbf{End-to-End Offline Evaluation}}
We conduct a comprehensive end-to-end evaluation of TURA against three strong baselines: LLM + RAG, Dynamic RAG, and Tool-Agent on the MCP-Bench dataset. Table~\ref{tab:overall_performance} shows that TURA achieves substantial improvements in both answer accuracy and faithfulness across human and automated evaluations.
\begin{table}[!t]
\centering
\caption{End-to-end performance comparison on MCP-Bench.}
\label{tab:overall_performance}
\vspace{-6pt}
\setlength\tabcolsep{3pt}
\renewcommand{\arraystretch}{1.12}
\small

\begin{tabular}{
L{0.24\linewidth} 
| P{0.14\linewidth} 
P{0.14\linewidth} 
| P{0.14\linewidth} 
P{0.14\linewidth}}
\toprule
\textbf{Method} 
& \multicolumn{2}{c|}{\textbf{Accuracy}} 
& \multicolumn{2}{c}{\textbf{Faithfulness}} \\
& \textbf{Human} & \textbf{LLM} & \textbf{Human} & \textbf{LLM} \\
\midrule
LLM + RAG & 65.3\% & 68.1\% & 72.4\% & 75.0\% \\
Dynamic RAG & 67.2\% & 69.5\% & 77.6\% & 79.4\% \\
Tool-Agent & 76.8\% & 80.4\% & 81.7\% & 83.9\% \\
\midrule
\rowcolor{gray!15}
\textbf{TURA} & \textbf{87.5\%} & \textbf{88.3\%} & \textbf{96.2\%} & \textbf{97.1\%} \\
\bottomrule
\end{tabular}
\vspace{-4mm}
\end{table}
TURA demonstrates significant performance gains over three strong baselines. In answer accuracy, TURA achieves 87.5\% versus RAG's 65.3\% in human evaluation. This substantial gain highlights the limitations of passive retrieval for complex multi-faceted queries and validates our hypothesis that active tool planning is essential for robust performance.
The improvement in faithfulness is even more pronounced. TURA achieves 96.2\% faithfulness compared to RAG's 72.4\% in human evaluation. This difference stems from a fundamental architectural advantage: while RAG relies on synthesis from potentially noisy text corpora and is prone to hallucination, TURA's framework enables dynamic invocation of verified tools that provide high-fidelity information.
The strong correlation between human and LLM evaluations across both methods validates the reliability of automated evaluation approaches for this task domain.

\subsubsection{\textbf{Online Deployment and A/B Testing Results}}
Following the promising offline result, TURA was deployed in a live A/B test against the incumbent LLM + RAG production system. 
We sampled 103 queries to balance coverage and cost, combining high-frequency production requests with random samples to approximate real user demand, while accounting for potential execution failures. This domain mix aligns well with MCP-Bench. We observe that our largest gains stem from travel-related scenarios, a domain characterizing real-time and constraint-heavy tasks such as itinerary planning and availability checks.

Each query is independently evaluated by trained annotators following a shared rubric. Particularly, we define \emph{accuracy} as whether the response correctly satisfies the user intent, \emph{content value} as the degree to which the response provides actionable and decision-supporting information,
and \emph{satisfaction} as the overall user-perceived utility considering correctness, completeness, and presentation. Each result is annotated by three annotators, and in cases of disagreement, the final decision is made through voting.
 
As shown in Table \ref{tab:online_ab_test}, TURA delivered statistically significant improvements across key business metrics, demonstrating consistent advantages in both session satisfaction and response quality distribution. The online results confirm TURA's superiority. It increased the Session Success Rate by 8.9\% and achieved an 8.7\% overall performance advantage. In head-to-head comparisons, TURA was rated as "Good" (strictly better than the baseline) in 13\% of cases, while maintaining "Satisfactory" performance in 86\% of cases and reducing "Bad" ratings to only 4\%.

\begin{table}[!t]
\centering
\caption{Online A/B testing results comparing TURA with the production baseline. GSB shows TURA's performance advantage over LLM + RAG baseline.}
\vspace{-6pt}
\setlength\tabcolsep{3pt}
\renewcommand{\arraystretch}{1.1}
\small
\begin{tabular}{L{0.36\columnwidth}| P{0.26\columnwidth}| P{0.3\columnwidth}}
\toprule
\textbf{Online System} 
& \textbf{SSR} 
& \textbf{GSB} \\
\midrule
LLM + RAG (Legacy) & 55.1\% & - \\
\rowcolor{gray!15}
\textbf{TURA (Ours)} & \textbf{64.0\% (+8.9\%)} & \textbf{13\% / 86\% / 4\%} \\
\bottomrule
\end{tabular}
\label{tab:online_ab_test}
\vspace{-3mm}
\end{table}

\begin{table}[!t]
\centering
\caption{Detailed performance analysis showing TURA's 8.7\% performance advantage with significantly reduced error rates across all issue categories.}
\vspace{-6pt}
\setlength\tabcolsep{3pt}
\renewcommand{\arraystretch}{1.1}
\small
\begin{tabular}{l | P{0.2\linewidth} | P{0.2\linewidth} | P{0.22\linewidth}}
\toprule
\textbf{Online System} 
& \textbf{Total Issues} 
& \textbf{Advantage Rate} 
& \textbf{SSR} \\
\midrule
LLM + RAG (Legacy) 
& 66 
& - 
& 55.1\% \\
\rowcolor{gray!15}
\textbf{TURA (Ours)} 
& \textbf{55 (-16.7\%)} 
& \textbf{8.7\%} 
& \textbf{64.0\% (+8.9\%)} \\
\bottomrule
\end{tabular}
\vspace{-4mm}
\label{tab:detailed_performance}
\end{table}

Table~\ref{tab:detailed_performance} presents the detailed issue statistics and comparative performance metrics between \method and the LLM + RAG (i.e., online legacy system). The comparative results revealed that TURA's tool-calling capabilities were a key performance driver, enabling it to excel in scenarios requiring real-time data accuracy where the LLM + RAG baseline failed. For instance, the baseline shows significant temperature deviations in weather queries and major discrepancies in train schedules, whereas TURA provided precise, up-to-date information directly from authoritative sources. This superiority translated to a sharp reduction in critical failures from 9 in the baseline to just 4 in TURA. Overall, TURA reduced the total issue count by 16.7\% (from 66 to 55), with consistent improvements across all categories: accuracy (-7.1\%), content richness (-28.6\%), and content value (-17.6\%), indicating a substantial enhancement in response informativeness and reliability.
Given its robust performance improvements and consistent advantages across multiple evaluation metrics, TURA has demonstrated clear superiority over traditional LLM + RAG baselines, validating the effectiveness of synergizing RAG with agentic tool-use for accessing both static and dynamic information in industrial AI search productions.

\subsection{Ablation Studies and Component Analysis}

\subsubsection{\textbf{Analysis of Intent-Aware MCP Server Retrieval (RQ2)}}

To investigate the efficacy of our retrieval module, we performed detailed ablations. As shown in Table~\ref{tab:retrieval_ablation}, both query decomposition and index augmentation are indispensable. Removing decomposition (\textbf{w/o Decomp.}) severely impairs performance, confirming that a single vector cannot handle multi-intent queries. Removing augmentation (\textbf{w/o Augment.}) also causes a significant drop, demonstrating its necessity in bridging the semantic gap between user queries and server documentation. The \textbf{full TURA} model, integrating both, dramatically outperforms all variants.

\begin{table}[!t]
\centering
\caption{Ablation study of the retrieval module on MCP-Bench. Both query decomposition and index augmentation are critical for performance.}
\label{tab:retrieval_ablation}
\vspace{-6pt}
\setlength\tabcolsep{3pt}
\renewcommand{\arraystretch}{1.1}
\small
\begin{tabular}{L{0.46\linewidth}| P{0.22\linewidth}| P{0.22\linewidth}} 
\toprule
\textbf{Retrieval Method} 
& \textbf{Recall@5} $\uparrow$ 
& \textbf{Precision@5} $\uparrow$ \\
\midrule
Dense Retrieval (ERNIE) & 0.4187 & 0.5505 \\
TURA (w/o Augment.)     & 0.7500 & 0.8555 \\
TURA (w/o Decomp.)      & 0.4530 & 0.5631 \\
\rowcolor{gray!15}
\textbf{TURA (Full)}    & \textbf{0.8289} & \textbf{0.9190} \\
\bottomrule
\end{tabular}
\vspace{-3mm}
\end{table}

\begin{figure}[ht]
    \centering
    \includegraphics[width=0.88\columnwidth]{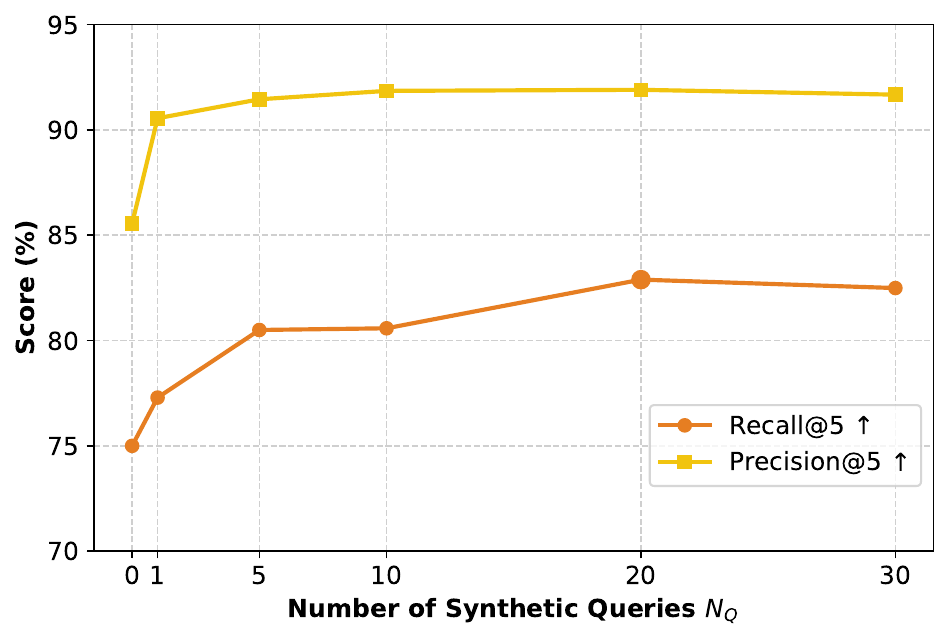}
    \caption{Impact of the number of synthetic queries per server on retrieval Recall@5. Performance peaks at 20 queries.}
    \vspace{-4mm}
    \label{fig:num_queries}
\end{figure}

\begin{table}[ht]
\centering
\caption{Comparison of different index representation strategies. Multi-vector approaches excel due to higher representational granularity. [$N_Q=20$]}
\label{tab:format_ablation}
\vspace{-6pt}
\setlength\tabcolsep{3pt}
\renewcommand{\arraystretch}{1.1}
\small
\begin{tabular}{L{0.50\linewidth}| P{0.20\linewidth}| P{0.20\linewidth}}
\toprule
\textbf{Index Representation Strategy} 
& \textbf{Recall@5} $\uparrow$ 
& \textbf{P@5} $\uparrow$ \\
\midrule
Single-Vector & 0.7721 & 0.8956 \\
\rowcolor{gray!15}
\textbf{Multi-Vector (Queries + Doc)} & \textbf{0.8289} & \textbf{0.9190} \\
Multi-Vector (Queries Only) & 0.8299 & 0.9208 \\
\bottomrule
\end{tabular}
\vspace{-4mm}
\end{table}

\begin{table}[htbp]
\centering
\caption{Impact of the DAG-based Task Planner on the complex, multi-hop query subset of MCP-Bench.}
\label{tab:planner_ablation}
\vspace{-6pt}
\setlength\tabcolsep{3pt}
\renewcommand{\arraystretch}{1.1}
\small
\begin{tabular}{L{0.32\linewidth}| P{0.26\linewidth}| P{0.32\linewidth}}
\toprule
\textbf{Planning Method} 
& \textbf{Success Rate (\%)} 
& \textbf{Avg. Latency (ms)} $\downarrow$ \\
\midrule
Sequential Plan & 88.9\% & 1,650 \\
\rowcolor{gray!15}
\textbf{DAG Plan} & \textbf{89.1\%} & \textbf{920} \\
\bottomrule
\end{tabular}
\vspace{-3mm}
\end{table}

\begin{table*}[ht!]
\centering
\caption{Performance of the distilled agent executor. Distillation significantly improves both accuracy and latency over the base models, achieving near-teacher accuracy at a fraction of the cost. P80 latency is reported.}
\vspace{-8pt}
\setlength\tabcolsep{4pt}
\renewcommand{\arraystretch}{1}
\small
\begin{tabular}{
L{0.24\textwidth} 
| P{0.12\textwidth} 
| P{0.22\textwidth} 
| P{0.22\textwidth}}
\toprule
\textbf{Agent Model} 
& \textbf{Model Size} 
& \textbf{Tool Calling Acc. (\%)} $\uparrow$ 
& \textbf{Avg. Latency/Step P80 (ms)} $\downarrow$ \\
\midrule
GPT-4o & N/A & 81.7 & 6,800 \\
Teacher (Deepseek-V3) & 671B-A37B & 82.4 & 8,700 \\
\midrule
Qwen3-1.7B & 1.7B & 43.5 & 1,500 \\
Qwen3-4B & 4B & 70.1 & 2,200 \\
Qwen3-30B-A3B & 30B-A3B & 76.3 & 2,600 \\
\midrule
\rowcolor{gray!15}
\textbf{Qwen3-1.7B Distilled} & \textbf{1.7B} & \textbf{77.6} & \textbf{620} \\
\rowcolor{gray!15}
\textbf{Qwen3-4B Distilled} & \textbf{4B} & \textbf{88.3} & \textbf{750} \\
\rowcolor{gray!15}
\textbf{Qwen3-30B-A3B Distilled} & \textbf{30B-A3B} & \textbf{88.7} & \textbf{760} \\
\bottomrule
\end{tabular}
\vspace{-3.5mm}
\label{tab:distillation_ablation}
\end{table*}

We then analyzed the configuration of the index augmentation. First, we determined the optimal number of synthetic queries ($N_Q$). As shown in Figure~\ref{fig:num_queries}, performance peaks at $N_Q=20$ and then plateaus. This suggests 20 queries provide sufficient semantic coverage without adding noise, so we fix $N_Q=20$.

Next, we explored how to structure these queries in the index (Table~\ref{tab:format_ablation}). A \textbf{Single-Vector} approach, which concatenates all text into one document for embedding, suffers from semantic dilution and performs the worst. In contrast, \textbf{Multi-Vector} approaches, which create separate embeddings for different parts of the server's information, achieve superior performance. This is because they offer higher representational granularity, providing focused semantic targets. While using only synthetic queries (\textbf{Queries Only}) performs marginally best, we chose the \textbf{Queries + Doc} strategy. This method retains the original server document as a ``safety net'', ensuring robustness for queries not covered by the synthetic data, a crucial feature for real-world deployment.

\subsubsection{\textbf{Importance of the DAG-based Task Planner (RQ3)}}
While Table \ref{tab:overall_performance} shows the latency impact of the DAG planner on the entire dataset, we conduct a targeted analysis on a challenging subset of MCP-Bench containing only complex, multi-hop queries where parallelism is possible. This isolates the planner's contribution to efficiency.
As Table \ref{tab:planner_ablation} illustrates, the DAG-based planner reduces average latency by \textbf{44.2\%} on these complex queries by identifying and executing independent sub-tasks in parallel. This substantial efficiency gain is achieved with no degradation in the execution success rate, confirming the effectiveness of our planner in optimizing complex workflows for online latency.

\subsubsection{\textbf{Effectiveness of Agent Distillation (RQ4)}}
\label{sec:rq4}

To address RQ4, we conduct a comprehensive evaluation of our proposed agent distillation methodology. The objective is to produce compact, low-latency student models that not only retain but ideally surpass the task-solving capabilities of the large teacher model. The performance of the distilled student models is benchmarked against their base versions, the teacher model (Deepseek-V3), and a strong proprietary baseline (GPT-4o)\citep{hurst2024gpt}, focusing on two key metrics: function-calling accuracy and P80 inference latency.

The empirical results, presented in Table \ref{tab:distillation_ablation}, unequivocally demonstrate the efficacy of our approach. We highlight two primary findings. First, our distilled models achieve a remarkable level of performance, surpassing even the powerful teacher model. Specifically, the \texttt{Qwen3-4B Distilled} and \texttt{Qwen3-30B-A3B Distilled} models attain accuracies of 88.3\% and 88.7\%, respectively. These results are substantially higher than both the 671B parameter teacher (82.4\%) and the formidable GPT-4o baseline (81.7\%). This phenomenon, where the student outperforms the teacher, validates the high quality of the synthetic trajectories generated by our data curation pipeline, which effectively filters noise and crystallizes optimal reasoning paths into a targeted training dataset.

Second, the distillation process yields significant improvements over the base models in both accuracy and efficiency. For instance, the \texttt{Qwen3-4B Distilled} model boosts accuracy by +18.2 absolute percentage points over its base counterpart (from 70.1\% to 88.3\%) while concurrently achieving a 66\% reduction in P80 latency (from 2,200ms to 750ms). This dual enhancement is a direct consequence of our "train-with-thought, infer-without-thought" paradigm. During training, this technique imbues the student model with the teacher's complex reasoning patterns. At inference, the student directly generates the concise, final action, minimizing token output and thus latency.

In selecting the final model for deployment, we considered the trade-offs between performance and operational cost. While the \texttt{Qwen3-30B-A3B Distilled} model, a Mixture-of-Experts (MoE) architecture\citep{jiang2024mixtral}, registered the highest accuracy, we opted for the \texttt{Qwen3-4B Distilled} model. The rationale is rooted in deployment feasibility and long-term cost-effectiveness. While the 3B-activated MoE model achieves similar inference performance to the 4B dense model, it requires dual-GPU L20 deployment due to its large total parameter size. The 4B model, however, can be efficiently served on a single GPU.  This makes the \texttt{Qwen3-4B Distilled} model the most pragmatic choice, offering a superior balance between accuracy and sustainable deployment costs. In summary, our agent distillation framework successfully forges agents that are smaller, faster, and more accurate, demonstrating a viable path for deploying powerful yet efficient agents in production systems.

\section{Conclusion}
This paper introduced TURA, a novel agentic framework designed to bridge the gap between traditional static RAG systems and the growing demand for dynamic, real-time information access in modern AI search. TURA overcomes passive retrieval's limitations through a cohesive three-stage architecture: Intent-Aware Retrieval for precise tool selection, DAG-based Task Planning for latency-optimized parallel execution, and an efficient Distilled Agent Executor. This empowers AI search to handle complex, multi-faceted queries previously intractable for conventional RAG systems.
Rigorous empirical evaluation, validated by a large-scale online A/B test in a production environment, confirms TURA's significant superiority. It markedly outperforms strong baselines, delivering substantial gains in answer accuracy and faithfulness, and a notable increase in Session Success Rate. This work presents a production-proven blueprint for the next generation of conversational AI, demonstrating a clear paradigm shift from passive information retrieval to active, tool-augmented systems. By enabling the seamless integration of heterogeneous, real-time data sources, TURA establishes a new benchmark for building robust and scalable industrial-grade AI search products.

\newpage
\bibliographystyle{ACM-Reference-Format}
\balance
\bibliography{main}

\appendix

\section{Prompt Templates}
This section provides illustrative examples of the prompt templates used in our methodology. For brevity and clarity, we have abstracted the core instructions and structures. These templates are designed to guide the LLM in performing specific sub-tasks within our framework.
\subsection{Tool Profile Generation}
\label{app:tool_gen}
\begin{description}
    \item[Instruction] As a tech expert, analyze a tool's document to generate diverse, practical example queries that showcase its core functions.
    
    \item[Input] Document: \texttt{\{doc\}}
    
    \item[Output] The expected output is a JSON array of strings, where each string is an example query.
        \begin{verbatim}
[
    "Example query demonstrating feature A",
    "Another query for a different use case",
    "Query with specific parameters mentioned 
    in the doc"
]
        \end{verbatim}
\end{description}

\subsection{Query Decomposition}
\label{app:query_decomp}
\begin{description}
    \item[Instruction] Deconstruct the user query into independent, atomic sub-tasks. Ensure full coverage of the original intent while maintaining independence between tasks.
    
    \item[Input] User Query: \texttt{\{query\}}
    
    \item[Output] The expected output is a JSON object containing a list of atomic sub-tasks. For example, if the input query is \textit{I need to book a flight to Shanghai for next week and find a good local restaurant there.}, the output would be:
        \begin{verbatim}
{
    "tasks": [
        "book flight to Shanghai for next week",
        "find recommended restaurants in Shanghai"
    ]
}
        \end{verbatim}
\end{description}

\subsection{Task Planning with DAG}
\label{app:plan}
\begin{description}
    \item[Instruction] Analyze the user query and decompose it into a Directed Acyclic Graph (DAG) of executable sub-tasks. Define each task and its dependencies to create an optimal execution plan.
    
    \item[Input] User Query: \texttt{\{query\}}
    
    \item[Output] The expected output is a JSON object defining tasks and their dependencies in a DAG structure:
        \begin{verbatim}
{
    "tasks": {
        "T1": "task description 1",
        "T2": "task description 2"
    },
    "dependency": [
        "T1->T2"
    ]
}
        \end{verbatim}
\end{description}

\subsection{Tool Execution}
\label{app:executor}
\subsubsection{Optimizing for Correctness}
\begin{description}
    \item[Instruction] Given a query and a set of available tools, generate a step-by-step reasoning trace. Accurately select the best tool, extract parameters, and verify the result before proceeding.
    
    \item[Input] 
        \begin{itemize}
            \item User Query: \texttt{\{query\}}
            \item Available Tools: \texttt{\{available\_tools\}}
        \end{itemize}

    \item[Output] The expected output is a JSON object representing the reasoning trace for a single step of execution:
        \begin{verbatim}
{
    "thought": "Reasoning for tool selection...",
    "action": {
        "tool": "<tool_name>",
        "params": { ... }
    },
    "observation": "Result from tool execution...",
    "next_step": "..." 
}
        \end{verbatim}
\end{description}

\subsubsection{Optimizing for Efficiency}
\begin{description}
    \item[Instruction] Given a query and a set of available tools, generate the most efficient tool-calling trace. Minimize redundant steps, prefer single-step resolutions, and terminate as soon as the answer is found.
    
    \item[Input] 
        \begin{itemize}
            \item User Query: \texttt{\{query\}}
            \item Available Tools: \texttt{\{available\_tools\}}
        \end{itemize}

    \item[Output] The expected output is a JSON object representing one step in the efficient tool-calling trace:
        \begin{verbatim}
{
    "step": 1,
    "tool": "<tool_name>",
    "params": { ... },
    "result": "<output>",
    "terminate": true
}
        \end{verbatim}
\end{description}
\end{document}